%% file: main.tex
\documentclass[10pt,twocolumn,letterpaper]{article}

\usepackage{cvpr}      

\usepackage{multirow}
\usepackage{tikz}
\usepackage{pifont}
\usepackage{algorithm}
\usepackage{algorithmic}

\usepackage[accsupp]{axessibility}  
\newcommand*{\affaddr}[1]{#1} 
\newcommand*{\affmark}[1][*]{\textsuperscript{#1}}

\usepackage[symbol]{footmisc}

\newcommand*\circled[1]{\tikz[baseline=(char.base)]{
            \node[shape=circle,draw,inner sep=1pt, scale=0.7] (char) {#1};}}


\definecolor{cvprblue}{rgb}{0.21,0.49,0.74}
\usepackage[pagebackref,breaklinks,colorlinks,allcolors=cvprblue]{hyperref}
\def\paperID{14980} 
\def\confName{CVPR}
\def\confYear{2025}

\title{Difference Inversion: Interpolate and Isolate the Difference with Token Consistency for Image Analogy Generation}

\author{%
\large
Hyunsoo Kim\affmark[1,]\affmark[2], Donghyun Kim$^\dag$\affmark[1], Suhyun Kim$^\dag$\affmark[3]\\
\normalsize{\affaddr{\affmark[1] Korea University}}\
\normalsize{\affaddr{\affmark[2] Korea Institute of Science and Technology}}\
\normalsize{\affaddr{\affmark[3] Kyung Hee University}}\\
{\tt\small \{climba,d\_kim\}@korea.ac.kr,dr.suhyun.kim@gmail.com}
}

\begin{document}
\twocolumn[{%
    \renewcommand\twocolumn[1][]{#1}%
    \vspace{-3em}
    \maketitle
    \begin{center}
        \centering
        \includegraphics[width=0.99\textwidth]{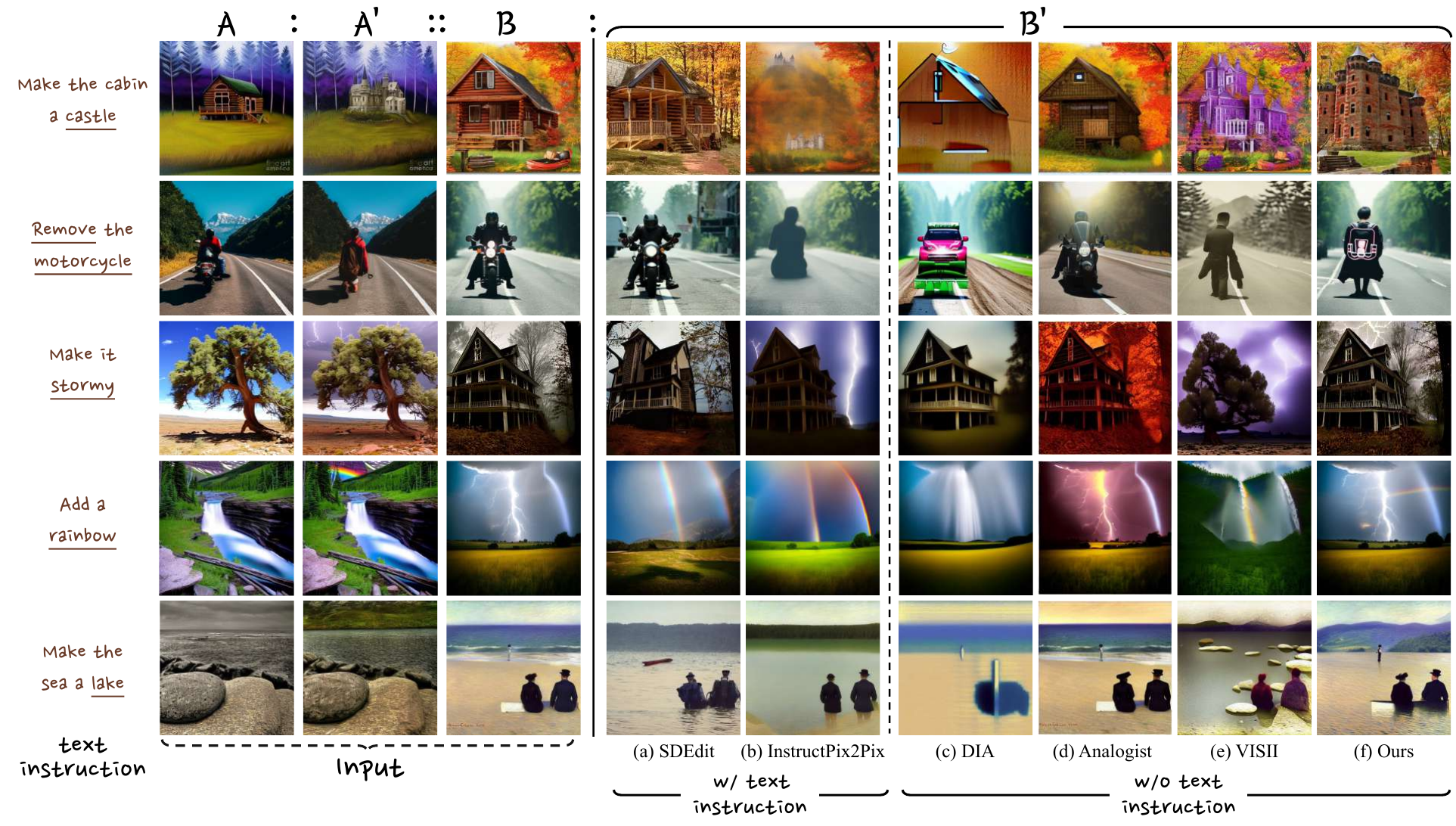}
        \captionof{figure}{
        We propose \textbf{\textit{Difference Inversion}}, a method that generates $B^\prime$ conditioned on an image triplet $\{A,A^\prime,B\}$ that satisfies $A:A^\prime::B:B^\prime$. Our method produces a significantly more plausible $B^\prime$ than other baselines. Note that SDEdit and InstructPix2Pix take the text instruction as input, whereas DIA, Analogist, VISII, and our Difference Inversion only use the image triplet ${A,A^\prime,B}$ as input.
        }
        \label{fig:teaser}
        \vspace{5pt}
    \end{center}%
    }]
    
\maketitle
\footnotetext[2]{Co-corresponding author.}
\input{sec/0_abstract}
\input{sec/1_intro}

\input{sec/2_relatedworks}
\input{sec/3_preliminaries}
\input{sec/4_method}
\input{sec/5_experiments}

\input{sec/6_limitation}
\input{sec/7_conclusion}
\input{sec/8_ack}
{
    \small
    \bibliographystyle{ieeenat_fullname}
    \bibliography{main}
}

\appendix

\end{document}

%% file: sec/0_abstract.tex
\begin{abstract}
How can we generate an image $B^\prime$ that satisfies $A:A^\prime::B:B^\prime$, given the input images $A$,$A^\prime$ and $B$?
Recent works have tackled this challenge through approaches like visual in-context learning or visual instruction. However, these methods are typically limited to specific models (\eg InstructPix2Pix. Inpainting models) rather than general diffusion models (\eg Stable Diffusion, SDXL). This dependency may lead to inherited biases or lower editing capabilities. In this paper, we propose Difference Inversion, a method 
that isolates only the difference from $A$ and $A^\prime$ and applies it to $B$ to generate a plausible $B^\prime$. To address model dependency, it is crucial to structure prompts in the form of a ``Full Prompt" suitable for input to stable diffusion models, rather than using an ``Instruction Prompt". To this end, we accurately extract the Difference between $A$ and $A^\prime$ and combine it with the prompt of $B$, enabling a plug-and-play application of the difference. To extract a precise difference, we first identify it through 1) Delta Interpolation. Additionally, to ensure accurate training, we propose the 2) Token Consistency Loss and 3) Zero Initialization of Token Embeddings. Our extensive experiments demonstrate that Difference Inversion outperforms existing baselines both quantitatively and qualitatively, indicating its ability to generate more feasible $B^\prime$ in a model-agnostic manner.
\end{abstract}

%% file: sec/1_intro.tex
\section{Introduction}
\label{sec:intro}
The goal of image analogy generation~\cite{vsubrtova2023diffusion} is to create the target image $B^\prime$ based on the image triplet \{$A$, $A^\prime$, $B$\}, ensuring that it satisfies the image analogy~\cite{hertzmann2023image, liao2017visual} formulation $A:A^\prime::B:B^\prime$. With the huge advancement of large-scale diffusion models~\cite{dhariwal2021diffusion, rombach2022high, saharia2022photorealistic, kawar2023imagic, podell2023sdxl, esser2024scaling}, it is natural to consider leveraging these models for solving image analogy generation from the perspective of conditional image generation. Unlike existing conditional diffusion models~\cite{batzolis2021conditional, meng2021sdedit, nichol2021glide, ho2022classifier, ramesh2022hierarchical, brooks2023instructpix2pix} that usually take text prompts as conditions, image analogy generation assumes that only an image triplet is provided as input. For example, consider the top example in Fig.~\ref{fig:teaser}, where the transformation is changing a cabin into a castle. Previous conditional diffusion models ((a) and (b)) use a text instruction (\ie ``Make the cabin a castle") as well as $B$ to generate the target image $B^\prime$. On the other hand, in image analogy generation, $B^\prime$ is generated using only the image triplet without any text prompt.

\begin{figure}[t!]
\centering
\includegraphics[width=1\linewidth]{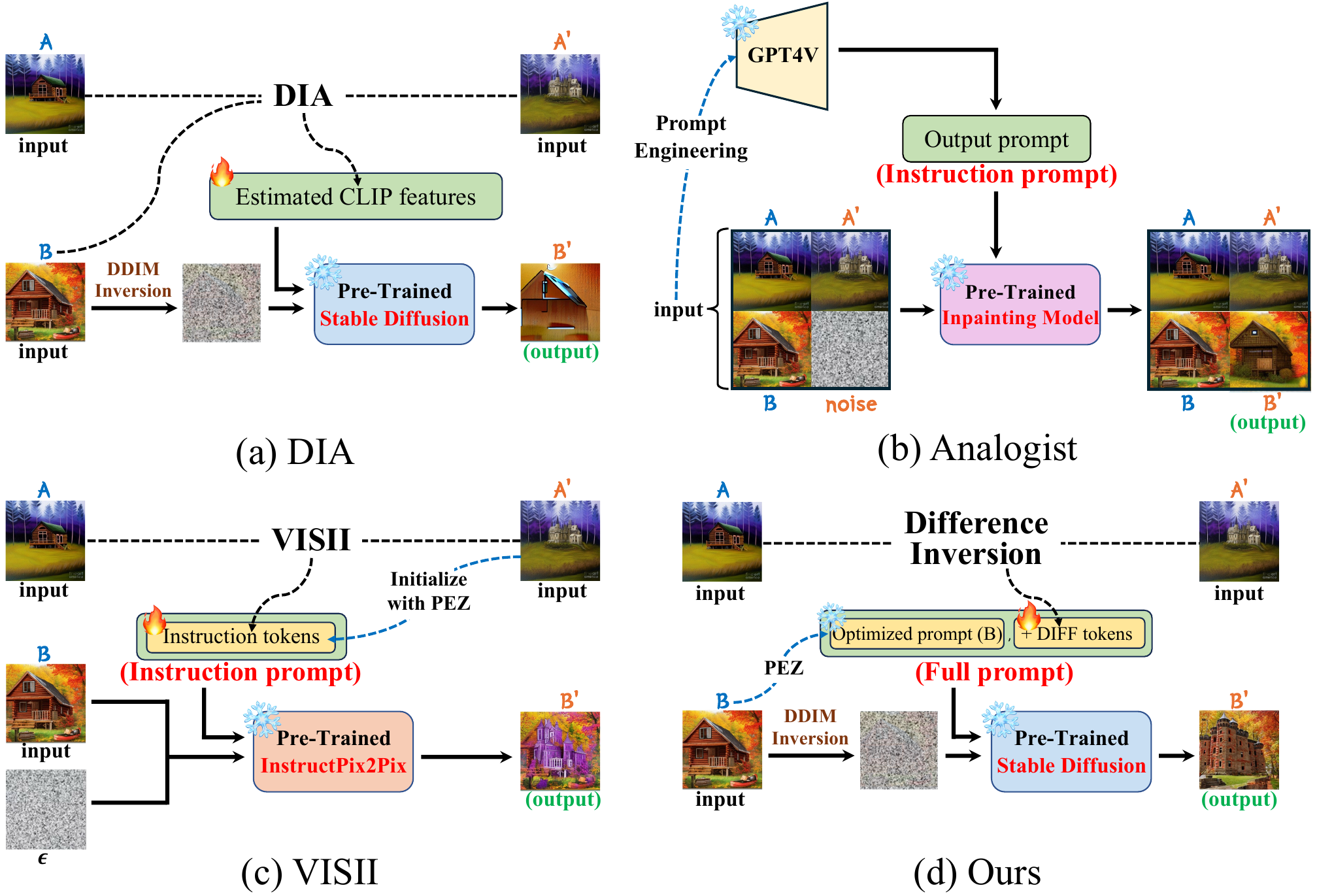}
\caption{\textbf{Architectural comparison of image analogy generation baselines.} 
We compare our approach with three baseline methods: (a) DIA, (b) Analogist, and (c) VISII. Detailed descriptions can be found in Sec.~\ref{sec:relatedworks}. It is noteworthy that, unlike (b) and (c), which each depend on specific models, our method can be applied to general stable diffusion models.}
\label{fig:model_comparison}
\end{figure}

Pioneered by DIA~\cite{vsubrtova2023diffusion}, several studies~\cite{vsubrtova2023diffusion,yang2024imagebrush,nguyen2024visual,gu2024analogist,meng2025instructgie} have explored diffusion-based image analogy generation in terms of visual in-context learning or visual instruction. From the perspective of visual in-context learning, ~\cite{yang2024imagebrush,gu2024analogist,meng2025instructgie} treat the generation of $B^\prime$ as an image inpainting task, where $A$, $A^\prime$, $B$ and initial noise are concatenated into a grid-based input image, and $B^\prime$ is denoised from noise (see Fig.~\ref{fig:model_comparison} (b)). This approach is limited to pretrained inpainting models and often requires manual prompt engineering by human to obtain additional text prompts. Visual Instruction Inversion (VISII, Fig.~\ref{fig:model_comparison} (c))~\cite{nguyen2024visual} inverts visual instructions from $A$ to $A^\prime$ using instruction tokens but is also limited to InstructPix2Pix~\cite{brooks2023instructpix2pix}. Moreover, to efficiently optimize the visual instruction, VISII initializes the instruction tokens with image $A^\prime$, which leaves residual information from $A^\prime$ in the instruction, leading to unintended artifacts in $B^\prime$. For instance, in the fourth example of Fig.~\ref{fig:teaser} (e), the transformation from $A$ to $A^\prime$ involves adding a rainbow, so the desired transformation for $B$ to $B^\prime$ would similarly involve adding only the rainbow. However, in the images generated by VISII, residual information from $A^\prime$ (\ie mountains) remains, leading to unintended elements in the output.

In this paper, we introduce Difference Inversion, which leverages ``Full Prompt'' with both prompt tokens representing the input image and Difference Tokens that exclusively encode difference information. Difference Tokens can focus on only the difference (Delta) between $A$ and $A^\prime$ and the combined tokens are more appropriate as an input of Stable Diffusion models.
This contrasts with VISII, which relies on an image-based ``Instruction Prompt" that can only be applied as input to InstructPix2Pix. To accurately disentangle the difference, we propose Delta Interpolation to define a more precise Delta. Additionally, we introduce Token Consistency Loss, which ensures that the difference information emerges when the token is present and disappears when it is absent. By using Token Consistency Loss, the Difference Tokens trained on the Interpolated Delta can be seamlessly concatenated with any query image prompt, enabling their use without introducing unwanted artifacts. Finally, we zero-initialize the Difference Tokens to ensure stable optimization of the differences and to reduce bias by preventing any initial preference from being introduced.

The main contributions of our paper are as follows:
\begin{itemize}
    \setlength\itemsep{0em}
    \item We propose Difference Inversion, which inverts precise differences into Difference Tokens and applies them to any query $B$ to generate $B^\prime$ within general stable diffusion frameworks.
    \item To extract the exact difference between images $A$ and $A^\prime$, we additionally introduce a Delta Interpolation, Token Consistency Loss and Zero-Initialization of Token Embeddings. These ensures that the extracted differences contain only the intended modifications, free from unwanted artifacts.
    \item Our extensive experiments demonstrate that Difference Inversion outperforms existing baselines both quantitatively and qualitatively, as well as in human and large-scale Vision Language Models (VLMs) evaluations.
\end{itemize}

%% file: sec/2_relatedworks.tex
\section{Related works}
\label{sec:relatedworks} 
\begin{figure*}[t!]
\centering
\includegraphics[width=1\linewidth]{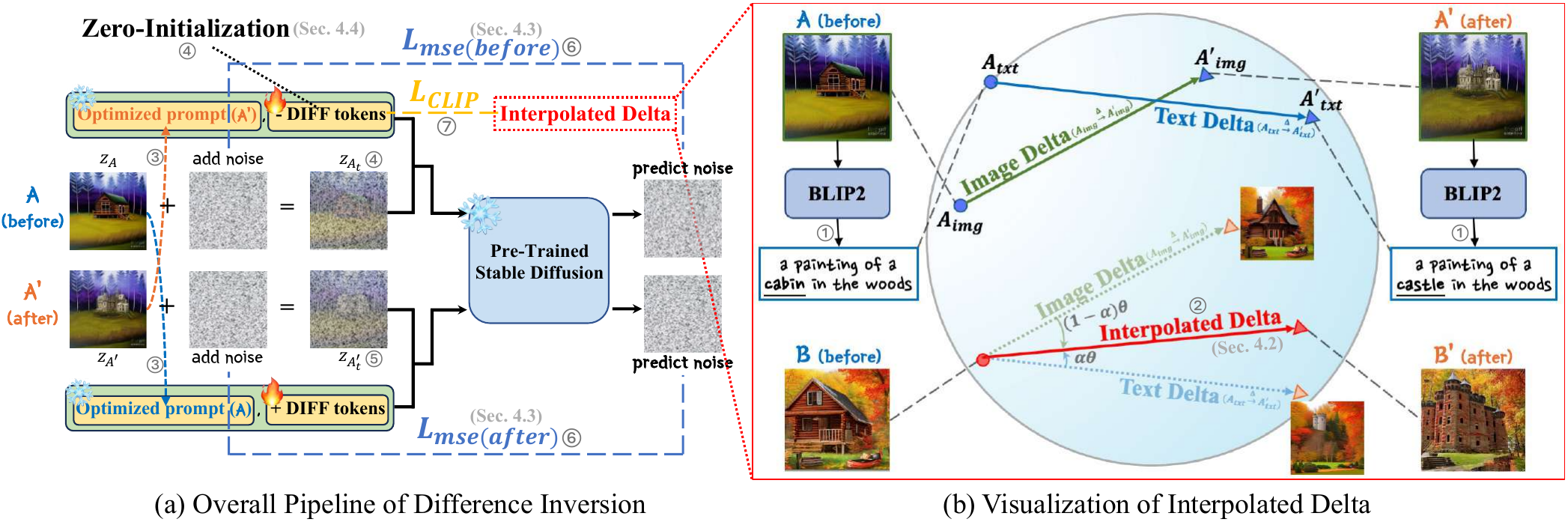}
\caption{\textbf{Overall pipeline of \emph{Difference Inversion}.} Given an image triplet \{$A,A^\prime,B$\}, Difference Inversion extracts the difference between $A$ and $A^\prime$ using DIFF tokens. The Difference is estimated as the Interpolated Delta, computed between A and A' by applying Spherical Linear Interpolation (Slerp) on the Image Delta and Text Delta, effectively capturing both visual and semantic information. The circled numbers in the figure correspond to the steps outlined in Algorithm 1. }
\label{fig:main_DIFF_inversion}
\end{figure*}

\subsection{Image Analogy Generation via Diffusion Models}
The goal of image analogy generation~\cite{vsubrtova2023diffusion,yang2024imagebrush,nguyen2024visual,gu2024analogist,meng2025instructgie} is to produce $B^\prime$ by conditioning on $A$, $A^\prime$ and $B$ in the form of $A:A^\prime::B:B^\prime$. DIA~\cite{vsubrtova2023diffusion} attempts to invert images to obtain conditioning matrices for diffusion, enabling high-level analogy through semantic vector operations. However it fails to capture not only the desired editing changes but also the detailed image information accurately. Starting with ImageBrush~\cite{yang2024imagebrush}, several studies combined $A$,$A^\prime$,$B$, and $B^\prime$ into a single grid-formatted image and used a pretrained inpainting model to denoise the $B^\prime$ section from noise, generating a plausible image for $B^\prime$. However, relying solely on a pretrained inpainting model makes it challenging to generate a plausible $B^\prime$. Leveraging the knowledge of large language models (LLMs) or large-scale vision language model (VLMs) is often necessary, which consequently requires cumbersome prompt engineering. In a different approach, VISII~\cite{nguyen2024visual} achieves $A \rightarrow A^\prime$ transformations by inverting visual instructions into text tokens and applying them to query image (\eg $B$) without relying on an VLMs. However, this process leaves unwanted information from $A$ and $A^\prime$ in $B^\prime$. As a result, $B^\prime$ may be generated in an unintended direction as shown in Fig.~\ref{fig:teaser} (c). Additionally, since the inverted instruction can only be used with InstructPix2Pix~\cite{brooks2023instructpix2pix}, it is challenging to apply this method to various Stable Diffusion models, resulting in low generalizability. In this paper, we introduce Difference Inversion that precisely extracts only the difference between $A$ and $A^\prime$ and applies it to $B$, avoiding unintended artifacts. Furthermore, since it forms a Full Prompt rather than an Instruction Prompt, it can be applied to a wide range of Stable Diffusion models.

\subsection{Diffusion based Inversion}
In the wide range of diffusion-based conditional generation methods, DDIM Inversion~\cite{song2020denoising,mokady2023null}, which identifies the initial noise corresponding to a given image, serves as a core component. Once the initial noise is identified by inversion, it is well-known that better results can be generated from the inverted noise when applied to personalized editing~\cite{kumari2023multi,gu2024photoswap,voynov2023p+} or image-to-image translation~\cite{hertz2022prompt,cheng2023general}. Alongside this trend, DreamBooth~\cite{ruiz2023dreambooth} and Textual Inversion~\cite{gal2022image} focus on optimizing special text tokens to learn representations that capture personalized information. ReVersion~\cite{huang2023reversion} also optimizes specific text prompts to capture and represent the relationship between two images. These inversion methods utilize a soft prompt approach, optimizing continuous embeddings. In contrast, PEZ~\cite{wen2024hard} proposes hard prompt optimization to generate interpretable prompts that effectively capture the concept of a specific image. Similarly, our work optimizes Difference Tokens in a hard prompt manner, making them plug-and-play with other prompts, thereby enhancing their suitability for general Stable Diffusion models.

%% file: sec/3_preliminaries.tex
\section{Preliminaries}
\label{sec:pre}
Denoising diffusion models~\cite{sohl2015deep, ho2020denoising} are generative models that learn the reversal of the forward process using denoising autoencoders, enabling the iterative sampling of images from noise. In the early stages of diffusion models~\cite{ho2020denoising, song2020score, dhariwal2021diffusion}, the forward and reverse diffusion process operates directly in the pixel space. Latent diffusion models~\cite{rombach2022high}, also well known as Stable Diffusion, performs this process in latent space, enabling more efficient image generation. Given an input image $x$ and a text condition $y$, the image encoder transforms $x$ into $z_0$. The denoising model $\epsilon_\theta$ is then trained to predict a denoised version of its input $z_t$, as shown in the following equation.
\begin{equation}
    \mathcal{L} = \mathbb{E}_{{\mathcal{E}(x), y, \epsilon \sim \mathcal{N}(0,1)}, t} \lVert \epsilon - \epsilon_{\theta}(z_{t}, t, \tau_\theta(y))\rVert^{2}_{2}
\end{equation}
where $z_t$ is a noisy variant of the original input $z_0$ in time t and $\tau_\theta$ is text encoder.
Since Stable Diffusion accepts text prompts as input, there have been attempts to add special tokens to text prompts for tasks such as personalized editing. Textual Inversion~\cite{gal2022image}, for example, learns to capture specific concepts or objects using special token ($s_*$).

Similarly, VISII (Fig.~\ref{fig:model_comparison} (c)) also utilizes instruction tokens $C_T$ to learn the instruction information necessary for generating $B^\prime$. Since VISII needs to learn the transformation instruction from image $A$ to $A^\prime$ and apply them to $B$, it leverages InstructPix2Pix~\cite{brooks2023instructpix2pix}. Furthermore, to better capture the transformation, $C_T$ is initialized via discrete token optimization (PEZ~\cite{wen2024hard}) on the image $A^\prime$. Since VISII relies on InstructPix2Pix, it learns $C_T$ in the form of an instruction prompt. In contrast, our Difference Inversion isolates only the difference information, enabling a plug-and-play approach with general stable diffusion prompts. Further details will be provided in the following section.

%% file: sec/4_method.tex
\section{Method}
\label{sec:method}

Given an image triplet $\{A, A^\prime,B\}$, our goal is to extract the difference between ($A, A^\prime$) and apply it to $B$. To capture the difference, we optimize DIFF tokens ($\tilde{D}$) to effectively encode the extracted differences. We first introduce the basic concept of Difference Inversion in Sec.~\ref{4.1}. In Sec.~\ref{4.2}, we explain how we optimize the DIFF token using the Interpolated Delta. Then, in Sec.~\ref{4.3}, we present the token consistency loss to enhance image consistency, followed by an introduction to token embedding initialization in Sec.~\ref{4.4}.

\subsection{Difference Inversion} \label{4.1}
We start from the key observation in VISII, which is the similar concept with our work. As shown in Fig.~\ref{fig:teaser} (e), generating $B^\prime$ with VISII results in unwanted artifacts. This occurs because the optimized visual instruction contains not only the difference (\eg rainbow) but also information specific to $A$ and $A^\prime$ (\eg mountains). Therefore, the key is to enable the inversion of only the precise difference.

In this paper, we define the difference as ``Delta," which can be obtained as the difference between embedding vectors from the CLIP~\cite{radford2021learning} encoder $\mathcal{E}_{I}$. Letting the CLIP image encoder be represented by $\mathcal{E}_{I}$, the Image Delta $\mathcal{D}_{img}$ can be expressed as follows:
\begin{equation}
\label{eq:TCloss}
\mathcal{D}_{img} = \mathcal{E}_{I}(A^\prime_{img}) - \mathcal{E}_{I}(A_{img})
\end{equation}

To optimize DIFF tokens, we first initialize the text prompts of $A$ and $A^\prime$ using PEZ. This step aims to fully anchoring the prompt of $A$ (or $A^\prime$) and isolate the only difference, enabling more precise extraction of the difference with the DIFF token in place. As shown in Fig.~\ref{fig:main_DIFF_inversion} (a), we only optimize DIFF tokens while optimized prompt embedding is frozen. Note that while optimization with PEZ can be time-consuming, it can be replaced with text captions generated by BLIP2~\cite{li2023blip}, trading off a slight performance drop for faster processing (Sec.~\ref{4.3})

However, simply optimizing the DIFF token only with the Image Delta did not sufficiently capture the Difference information. As shown in Fig.~\ref{fig:main_DIFF_inversion} (b), $B^\prime$ with the Image Delta retains much of the original image's cabin features rather than fully transforming into the intended castle. We hypothesize that this is because the CLIP embedding space may not fully represent the intricate details of the images. In the following section, we introduce Delta Interpolation to extract a more refined difference.

\begin{algorithm}[tb]
   \caption{Difference Inversion}
   \label{alg:example}
\begin{algorithmic}
   \STATE {\bfseries Require:} pretrained denoising model $\epsilon_\theta$; Image encoder $\mathcal{E}$;\ CLIP Image, Text Encoder $\mathcal{E}_{I}$, $\mathcal{E}_{T}$;\ Captioning model BLIP2;\ interpolation rate $\alpha$;\ Learning rate $\gamma$;\ $t_{before}, t_{after} \sim \mathcal{U}(0, T)$;\ $\epsilon_{before}, \epsilon_{after} \sim \mathcal{N}(0,1)$;\
   \STATE {\bfseries Input:} Image pair $\{A, A^\prime\}$
   \STATE {\bfseries Output:} Difference Tokens $\tilde{D} = \{D_1^*, ..., D_n^*\}$
   \STATE \textcolor{gray}{// Extract the caption of $A$ and $A^\prime$ \ldots \circled{1}}
   \STATE $cap_{A}$ = BLIP2($A$);\ $cap_{A^\prime}$ = BLIP2($A^\prime$)
   \STATE \textcolor{gray}{// Calculate Image Delta $\mathcal{D}_{I}$ and Text Delta $\mathcal{D}_{T}$}
   \STATE ${\mathcal{D}}_{I} = \mathcal{E}_{I}(A^\prime) - \mathcal{E}_{I}(A)$;\ ${\mathcal{D}}_{T} = \mathcal{E}_{T}(cap_{A^\prime}) - \mathcal{E}_{txt}(cap_{A})$
   \STATE \textcolor{gray}{// Calculate Interpolated Delta $\mathcal{D}_{inter}$ \ldots \circled{2}}
   \STATE ${\mathcal{D}}_{inter} = Slerp({\mathcal{D}}_{I}, {\mathcal{D}}_{T}, \alpha)$
   \STATE \textcolor{gray}{// Optimize the prompt w.r.t. $A$ and $A^\prime$ \ldots \circled{3}}
   \STATE $prompt_{A}$ = PEZ($A$);\ $prompt_{A^\prime}$ = PEZ($A^\prime$)
   \STATE Encode $z_{A} = \mathcal{E}(A)$;\ $z_{A^\prime} = \mathcal{E}(A^\prime)$
   \STATE \textcolor{gray}{// Zero-Initialization token embeddings \ldots \circled{4}}
   \STATE Initialize $\tilde{D}$ with Zero embeddings
   \FOR{$i=1$ {\bfseries to} $N$}
   \STATE \textcolor{gray}{// Prepare noisy latents $z_{A}$ and $z_{A^\prime}$ \ldots \circled{5}}
   \STATE $z_{A_{t}} \leftarrow$ add $\epsilon_{before}$ to $z_{A}$ at timestep $t_{before}$
   \STATE $z_{A^\prime_{t}} \leftarrow$ add $\epsilon_{after}$ to $z_{A^\prime}$ at timestep $t_{after}$
   \STATE \textcolor{gray}{// Predict each noise $\hat{\epsilon}_{before}, \hat{\epsilon}_{after}$}
   \STATE $\hat{\epsilon}_{before} = \epsilon_{\theta}(z_{A_t}, t_{before}, \{prompt_{A^\prime}, -\tilde{D}\})$
   \STATE $\hat{\epsilon}_{after} = \epsilon_{\theta}(z_{A^\prime_t}, t_{after}, \{prompt_{A}, \tilde{D}\})$
   \STATE \textcolor{gray}{// Compute Token Consistency loss \ldots \circled{6}}
   \STATE $\mathcal{L}_{tc} = \lVert \epsilon_{before} - \hat{\epsilon}_{before}\rVert_{2} + \lVert \epsilon_{after} - \hat{\epsilon}_{after}\rVert_{2}$
   \STATE \textcolor{gray}{// Compute Clip loss $\mathcal{L}_{clip}$ \ldots \circled{7}}
   \STATE $\mathcal{L}_{clip} = \text{cosine}(\tilde{D}, {\mathcal{D}}_{inter})$
   \STATE $\mathcal{L} = \lambda_{\textit{tc}} * \mathcal{L}_{tc} + \lambda_{clip} * \mathcal{L}_{clip}$
   \STATE Update $\tilde{D} = \tilde{D} - \gamma \nabla\mathcal{L}$
   \ENDFOR
   \STATE {\bfseries Return:} $\tilde{D}$
\end{algorithmic}
\end{algorithm}

\subsection{Delta Interpolation} \label{4.2}
Although the Image Delta effectively captures the visual difference, it still fails to fully reflect the specific semantic differences. To obtain more detailed semantic information, we use an image captioning model to extract captions for each of the images $A$ and $A^\prime$. We utilize BLIP2, but for more detailed captions, BLIP3~\cite{xue2024xgen} or large-scale VLMs~\cite{liu2024visual,liu2024llava, chen2024internvl} can be also be considered. The extracted captions contain more specific details, such as ``cabin" and ``castle," which were used to derive the Text Delta $D_{txt}$ that contains more semantic information than Image Delta.
\begin{equation}
\label{eq:TCloss}
\mathcal{D}_{txt} = \mathcal{E}_{T}(A^\prime_{txt}) - \mathcal{E}_{T}(A_{txt})
\end{equation}
where $\mathcal{E}_{T}$ is CLIP text encoder.

After obtaining both Image and Text Delta of $A$ and $A^\prime$ with CLIP, we apply Spherical Linear Interpolation (Slerp)~\cite{shoemake1985animating} to integrate both embeddings. Since CLIP learns image-text pairs to maximize their cosine similarity, we hypothesize that it would be possible to find a more refined Delta within the joint hypersphere of the CLIP space. Therefore, to identify the Interpolated Delta $\mathcal{D}_{inter}$ on this hypersphere, we apply Slerp shown below.
\begin{align}
\label{eq:SLERP}
\mathcal{D}_{inter} & = Slerp(\mathcal{D}_{img},\mathcal{D}_{txt};\alpha) \\ & = \frac{sin(({1-\alpha})\theta)}{sin(\theta)} \mathcal{D}_{img} +  \frac{sin(\alpha\theta)}{sin(\theta)} \mathcal{D}_{txt}
\end{align}
where $\alpha$ is interpolation ratio between Image and Text Delta. We set $\alpha$ to 0.8, and the ablation results are in Fig.~\ref{alpha_ablation}.

Finally, we optimize the DIFF token to minimize cosine distance with this Interpolated Delta as:
\begin{equation}
\label{eq:loss_clip}
    \mathcal{L}_{clip} = \text{cosine}(\tilde{D}, \mathcal{D}_{inter})
\end{equation}

\subsection{Token Consistency Loss} \label{4.3}
Using only the Interpolated Delta $\mathcal{D}_{inter}$ may resemble obtaining a visual instruction from $A$ to $A^\prime$. However, in Difference Inversion, it is also important to preserve aspects that should remain unchanged when generating $B^\prime$. To achieve this, it is essential to accurately extract only the bidirectional difference between $A$ and $A^\prime$. Visual instruction considers only the transformation from $A \rightarrow A^\prime$, whereas the difference must account for both $A \rightarrow A^\prime$ and $A^\prime \rightarrow A$. For DIFF tokens ($\tilde{D}$), this implies that when $\tilde{D}$ is present, the result should correspond to $A^\prime$; otherwise, it should yield $A$.

Additionally, maintaining consistency in elements we do not intend to modify during this process is necessary. Inspired by Cycle Consistency Loss~\cite{zhu2017unpaired, xu2024cyclenet}, we introduce Token Consistency Loss, which ensures that $A^\prime$ is generated when $\tilde{D}$ is present and $A$ when $\tilde{D}$ is absent. Specifically, the prompt concatenated with prompt $A$ and $\tilde{D}$ should accurately reconstruct image $A$, while the prompt concatenated with prompt $A^\prime$ and $- \tilde{D}$ (reverse direction of $\tilde{D}$) should accurately reconstruct image $A^\prime$. This can be expressed as follows:
\begin{equation}
\label{eq:MSE_before}
\mathcal{L}_{mse_{before}} = \lVert \epsilon_{before} - \epsilon_{\theta}(z_{A_t}, t_{before}, \{prompt_{A^\prime}, -\tilde{D}\})\rVert_{2}
\end{equation}
\begin{equation}
\label{eq:MSE_after}
\mathcal{L}_{mse_{after}} = \lVert \epsilon_{after} - \epsilon_{\theta}(z_{A^\prime_t}, t_{after}, \{prompt_{A}, \tilde{D}\})\rVert_{2}
\end{equation}
where $t_{before},t_{after} \sim N(0,1)$

The Token Consistency loss can be written as
\begin{equation}
\label{eq:TCloss}
\mathcal{L}_{tc} = \mathcal{L}_{mse_{before}} + \mathcal{L}_{mse_{after}}   
\end{equation}

Finally, to optimize the DIFF tokens $\tilde D$, we utilized the following objective:
\begin{equation}
\label{eq:final}
\mathcal{L} = \lambda_{tc} * \mathcal{L}_{tc} + \lambda_{clip} * \mathcal{L}_{clip}
\end{equation}
where $\lambda_{tc}$ and $\lambda_{tc}$ denotes weight parameter of each loss, respectively. We empirically set $\lambda_{tc}$ to 0.01 and $\lambda_{clip}$ to 6.

\subsection{Zero-Initialization of Token Embeddings} \label{4.4}
Thus far, we have discussed the optimization process for the DIFF tokens ($\tilde{D}$). Finally, we address the approach for initializing $\tilde{D}$. It is crucial for the $\tilde{D}$ to capture accurate differences, making it essential to effectively disentangle the original image information from the difference. From this perspective, randomly initializing the $\tilde{D}$ can lead to unintended information being embedded within $\tilde{D}$ (see Config. C in Fig.~\ref{fig:qualitative_ablation}). Motivated by ContorlNet~\cite{zhang2023adding}, we initialize the token embeddings with zero to ensure that only the intended differences are captured within the $\tilde{D}$, minimizing unwanted artifacts. By initializing all $\tilde{D}$ to zero, we ensure that each token robustly encapsulates only the distinguishing differences from the original image. Detailed experimental results are presented in Sec.~\ref{sec:experiments}.

%% file: sec/5_experiments.tex
\begin{figure*}[t!]
\centering
\includegraphics[width=0.8\linewidth]{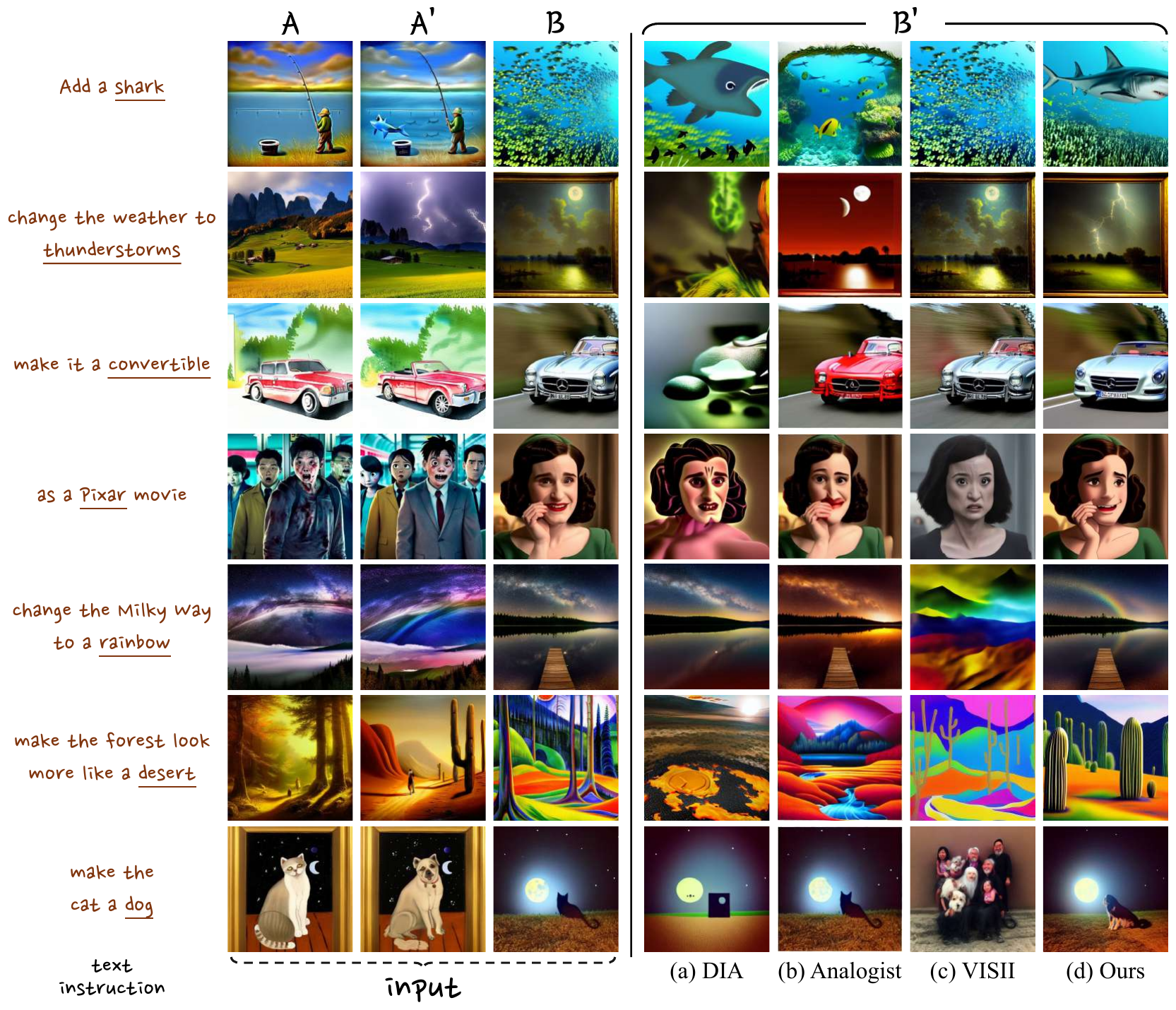}
\caption{\textbf{Qualitative comparison to baseline methods.} Given inputs $A$, $A^\prime$ and B, we generate $B^\prime$ without utilizing any text instructions.}
\label{fig:qualitative_main}
\end{figure*}

\section{Experiments}
\label{sec:experiments}

\subsection{Experimental and Implementation Detail} \label{5.1}
We used Stable Diffusion 2.1 as our baseline. All experiments were conducted on a single NVIDIA GeForce RTX 4090 GPU. For further details on the experimental setup, please refer to the supplementary materials.

\paragraph{Baselines.}
We compared our methodology with four baselines on image analogy generation. These methods all generate $B^\prime$ using only an image triplet \{$A, A^\prime, B$\}. Note that, some grid-image-based approaches (ImageBrush~\cite{yang2024imagebrush} and InstructGIE~\cite{meng2025instructgie}) are also close with our works but since their code is not publicly available, we did not include them in the comparison. The implementation details for each baseline are provided in the supplementary materials.

\paragraph{Dataset.}
We randomly sample 300 instruct pairs from the InstructPix2Pix dataset to measure quantitative and qualitative results. The InstructPix2Pix dataset is structured as triplets in the form of \{before image, after image, text instruction\}. We utilized the before image and after image as $A$ and $A^\prime$, respectively. $B$ was randomly selected from other before images with the same text instruction. Note that we do not use the text instruction included in the dataset.

\paragraph{Evaluation Metric.}
We evaluate how well the difference is reflected using the directional score using CLIP. We also measure the directional score using DINO-v2, which offers a more refined embedding space than CLIP. Additionally, we conduct a user study to verify the perceptual suitability of our results from a human perspective. And finally, leveraging the significant advancements in large-scale vision language models (VLMs), we perform prompt engineering to enable VLMs to evaluate our tasks, as shown in the figure 5. We adopt GPT-4o~\cite{OpenAI_2022}, Qwen2-72B~\cite{wang2024qwen2} and Llama3.2-90B~\cite{dubey2024llama} for VLM evaluation, and the question prompts are provided in the supplementary materials. For human evaluation, we use a four-option multiple-choice format with randomized order of the options. On the other hand, for VLM evaluation, we employ a two-option format to minimize the influence of numbering and enable a more precise comparison. Further details about human and LLM evaluations are in the supplementary materials.

\subsection{Qualitative Results}
As shown in Fig.~\ref{fig:qualitative_main}, our approach qualitatively generates significantly superior images compared to existing baselines. It not only accurately reflects the transformation from $A \rightarrow A^\prime$ but also preserves as much information from B as possible, resulting in a plausible $B^\prime$. In Fig.~\ref{fig:teaser}, (a), (b), and (c) depict results generated using only the images $A$, $A^\prime$, and $B$ without text instructions, while (d) and (e) utilize text instructions. Across all baselines, our results produce a $B^\prime$ that best captures the before-after differences.

\begin{table}[t]
\centering
\resizebox{1.0\linewidth}{!} {%
\begin{tabular}{c|cc|cccc}
\toprule
\multirow{2}{*}{\textbf{Metric}} & \multicolumn{2}{c|}{\textbf{Image + Text}} & \multicolumn{4}{c}{\textbf{Image only}} \\ 
\cmidrule{2-7} 
 & SDEdit & InstructPix2Pix & DIA & Analogist & VISII & Ours \\ 
\midrule
CLIP ($\uparrow$) & 0.1105 & \textbf{0.2000} & 0.0294 & 0.0398 & \underline{0.1007} & \textbf{0.1024} \\
DINO-v2 ($\uparrow$) & 0.5374 & \textbf{0.5723} & 0.5114 & 0.5098 & \underline{0.5414} & \textbf{0.5732} \\
\bottomrule
\end{tabular}}
\caption{\textbf{Directional Score of CLIP and DINO-v2.} We conducted quantitative evaluation by measuring the directional score using CLIP and DINO-v2. The directional score represents the embedding similarity between $A \rightarrow A^\prime$ and $B \rightarrow B^\prime$.}
\label{table:1}
\end{table}

\begin{figure}[t!]
\centering
\includegraphics[width=1\linewidth]{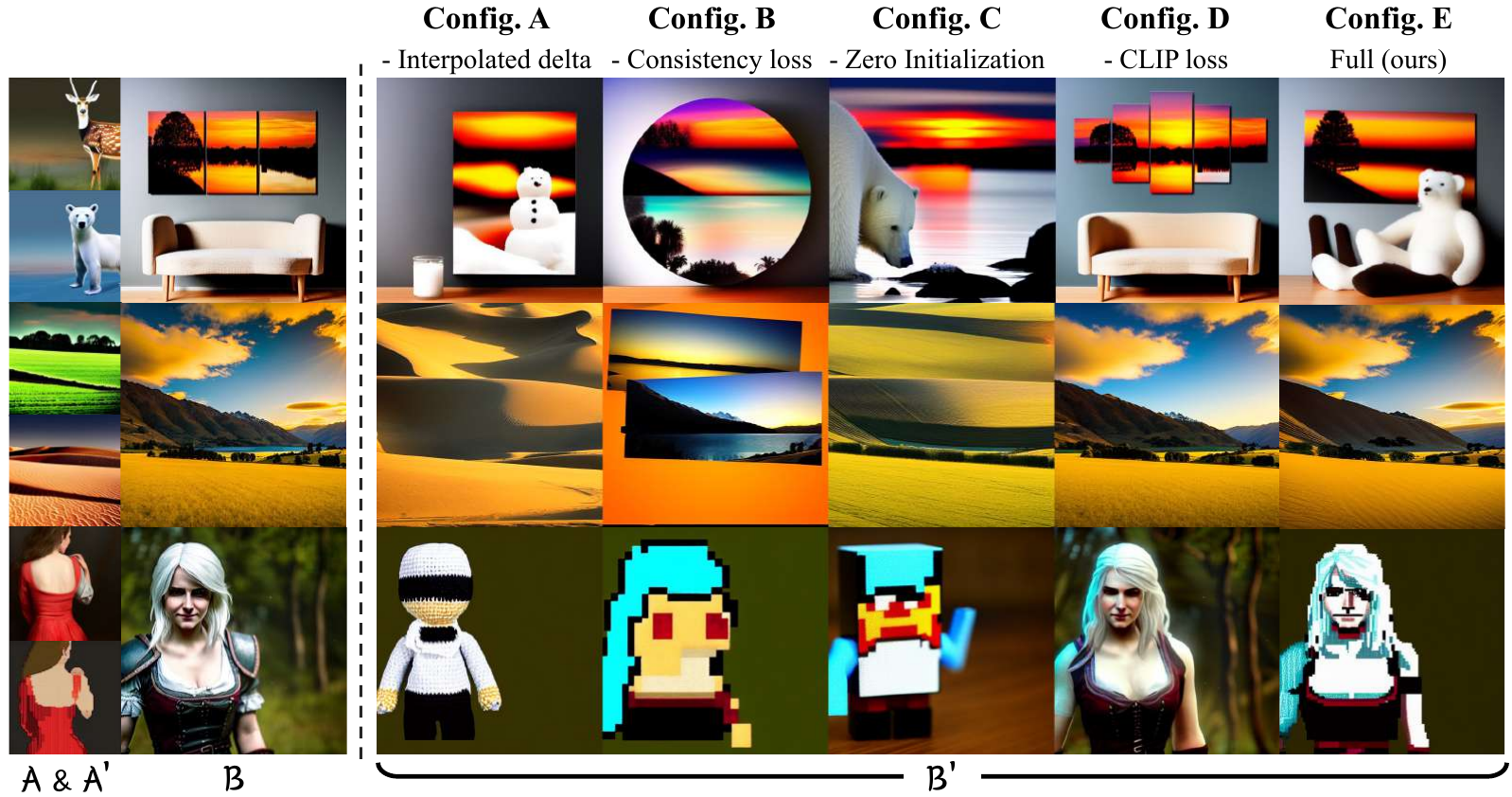}
\caption{\textbf{Qualitative comparison with ablation studies for each proposed method.} Each column shows the results when the corresponding method \textbf{\textit{is not applied.}}}
\label{fig:qualitative_ablation}

\end{figure}

\subsection{Quantitative Results}
\paragraph{Directional score.}
As shown in Table~\ref{table:1}, our model achieves the best performance among image-only baselines in terms of the directional score using CLIP. Note that VISII's CLIP score is nearly comparable to ours, which is attributable to the fact that the dataset used for evaluation was also utilized in training VISII's backbone, InstructPix2Pix. In other words, VISII operates on a dataset it has already been trained on, whereas our model is evaluated on previously unseen data. Despite using Stable Diffusion as backbone, our model still performed exceptionally well. This gap becomes even clearer when measured with DINO-v2~\cite{oquab2023dinov2} rather than CLIP. Furthermore, our DINO-v2 score surpasses even SDEdit and InstructPix2Pix, both of which utilize text descriptions.

\begin{table}[t]
\centering
\resizebox{0.9\linewidth}{!} {%
\begin{tabular}{c|c|c|c|c}
\toprule
\textbf{Model} & DIA & Analogist & VISII & Ours \\
\midrule
\textbf{Percentage} (\%) & 4\% & 15\% & 9\% & 72\% \\
\bottomrule
\end{tabular}}
\caption{\textbf{Human evaluation for image analogy generation}. In each question, participants were asked which $B^\prime$ would best complete the analogy $A:A^\prime::B:B^\prime$. We report the percentage preference for Difference Inversion over 50 images evaluated by 60 people.}
\label{table:2}
\end{table}

\paragraph{Human evaluation.}
Although DINO-v2 offers a better directional score compared to CLIP, it still does not fully capture how plausible $B^\prime$ is in the $A:A^\prime :: B:B^\prime$ relationship. To address this, we conduct human evaluations to assess which $B^\prime$ generated by the various baselines was the most appropriate. We asked 60 participants to select the most suitable image analogy generation result ($B^\prime$) from four baselines, including our method. The order of models is randomly assigned for each evaluation, and the images used in the survey are available in the supplementary materials. As shown in Table~\ref{table:2}, our model significantly outperforms the other three baselines in the human evaluation.

\begin{table}[t]
\centering
\resizebox{0.9\linewidth}{!} {%
\begin{tabular}{c|c|c|c}
\toprule
\textbf{Model (vs Ours)} & \textbf{DIA} & \textbf{Analogist} & \textbf{VISII} \\
\midrule
GPT4o~\cite{OpenAI_2022} & 90\% & 76\% & 74\% \\
Qwen2-VL-72B~\cite{wang2024qwen2} & 92\% & 70\% & 80\% \\
Llama-3.2-90B~\cite{dubey2024llama} & 72\% & 60\% & 62\% \\
\bottomrule
\end{tabular}}

\caption{\textbf{VLM evaluation for image analogy generation.} We provide the large-scale VLM with a prompt explaining the image analogy generation task and ask it to select the most suitable image for $B^\prime$, similar to the human evaluation (see in Sec.~\ref{5.1}). The results above reflect the win rate of Difference Inversion compared to each baseline in pairwise comparisons, evaluated using 50 examples for each model (a total of 150 evaluations each).}
\label{tabel:3}
\end{table}

\paragraph{VLM evaluation.}
We also conduct evaluation with large-scale VLMs known for their strong reasoning capabilities. We leverage GPT-4o~\footnote{\url{https://chatgpt.com/}}, Qwen2-VL-72B-Instruct~\footnote{\url{https://huggingface.co/Qwen/Qwen2.5-72B-Instruct}}, and Llama-3.2-90B-Vision~\footnote{\url{https://huggingface.co/meta-llama/Llama-3.2-90B-Vision}} for evaluation, and unlike human evaluation, we employed a two-option format to compare our model against the baselines. As shown in Table~\ref{tabel:3}, the preference for our model is significantly higher compared to existing baselines. This indicates that our method not only extracts the exact difference but also can generates plausible $B^\prime$, fulfilling the primary goal of image analogy generation.

\begin{figure}[t!]
\centering
\includegraphics[width=1\linewidth]{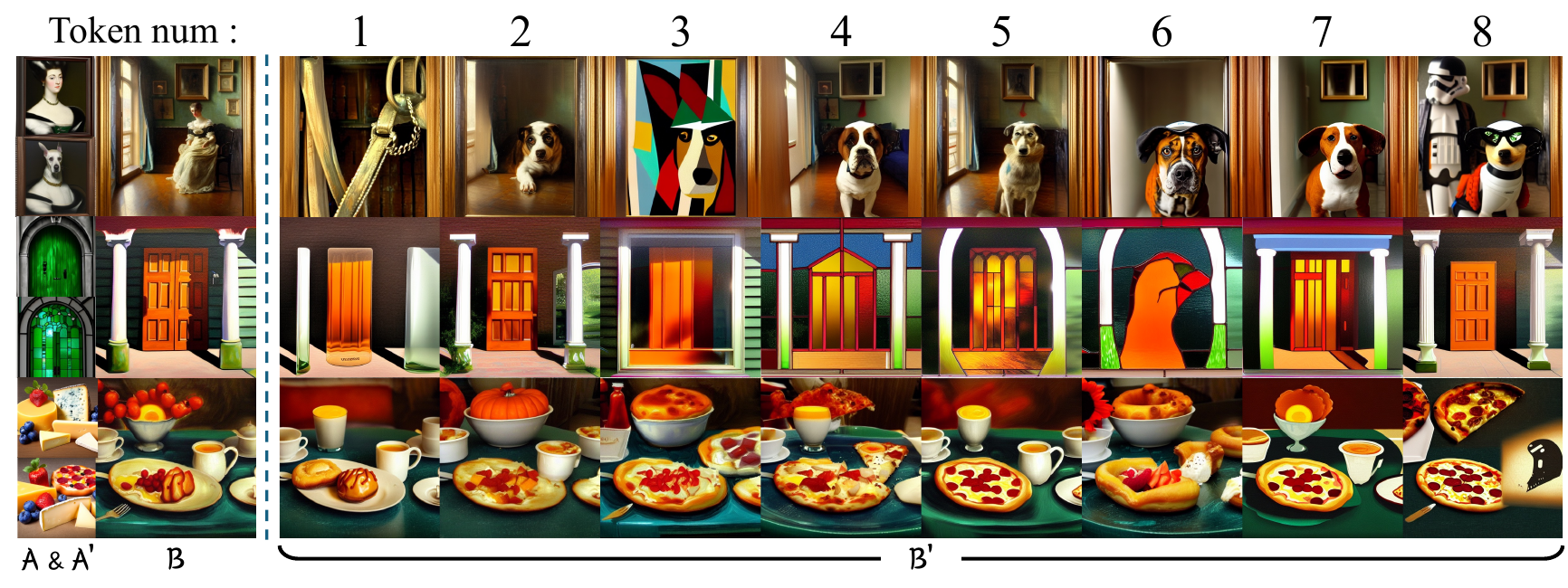}
\caption{\textbf{Ablation results on the number of Difference tokens ($\Tilde{D}$).} The fewer $\Tilde{D}$ used, the less effectively they capture the difference information; however, using too many $\Tilde{D}$ can lead to capturing additional, unrelated information. Experimentally, we found that using a total of 5 tokens strikes the right balance.}
\label{token_num}
\end{figure}

\begin{figure}[t!]
\centering
\includegraphics[width=1\linewidth]{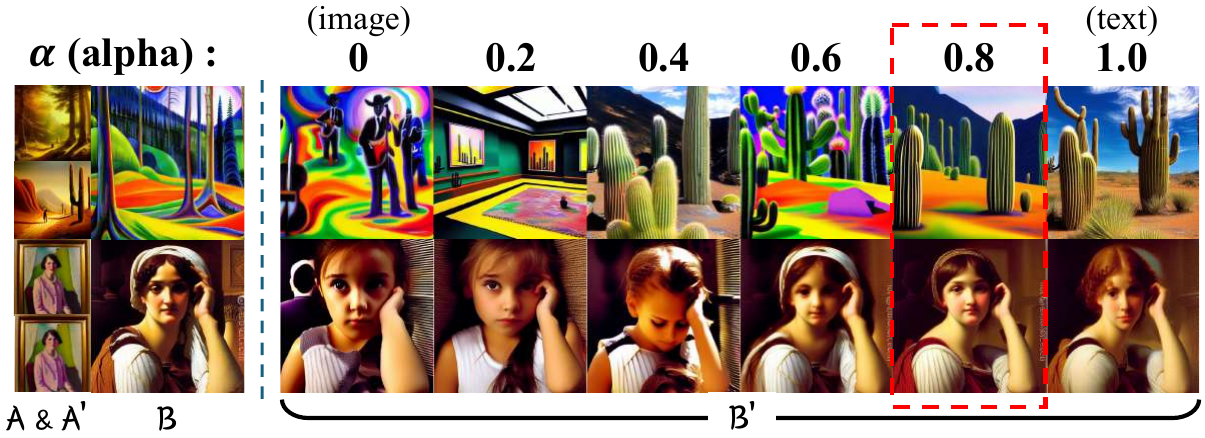}
\caption{\textbf{Ablation results of $\alpha$.} Since our delta ultimately needs to be optimized into text tokens, assigning a higher weight to $\alpha$ tends to better capture the desired differences. We finally set $\alpha$ to 0.8 for all experiments.}
\label{alpha_ablation}
\end{figure}

\subsection{Further Analysis}
\paragraph{Ablation study.}

Lastly, we performed ablation experiments on each of our proposed methods. Fig.~\ref{fig:qualitative_ablation} presents the ablation results for Token Consistency Loss, Delta Interpolation, Zero Initialization and CLIP loss. When Token Consistency Loss is not used, we observed significant degradation in consistency with the original image (B). Without Delta Interpolation and using only Image Delta, the method fails to capture the difference effectively. Similarly, Zero Initialization of Token Embedding shows that when initialized randomly, the difference information is not properly reflected. The ablation results for the number of difference tokens and the interpolation ratio alpha are shown in Fig.~\ref{token_num} and ~\ref{alpha_ablation}, respectively. Additional hyperparameter search results for each method, including lambda tc and lambda clip, can be found in the supplementary materials.

\begin{figure}[t!]
\centering
\includegraphics[width=1\linewidth]{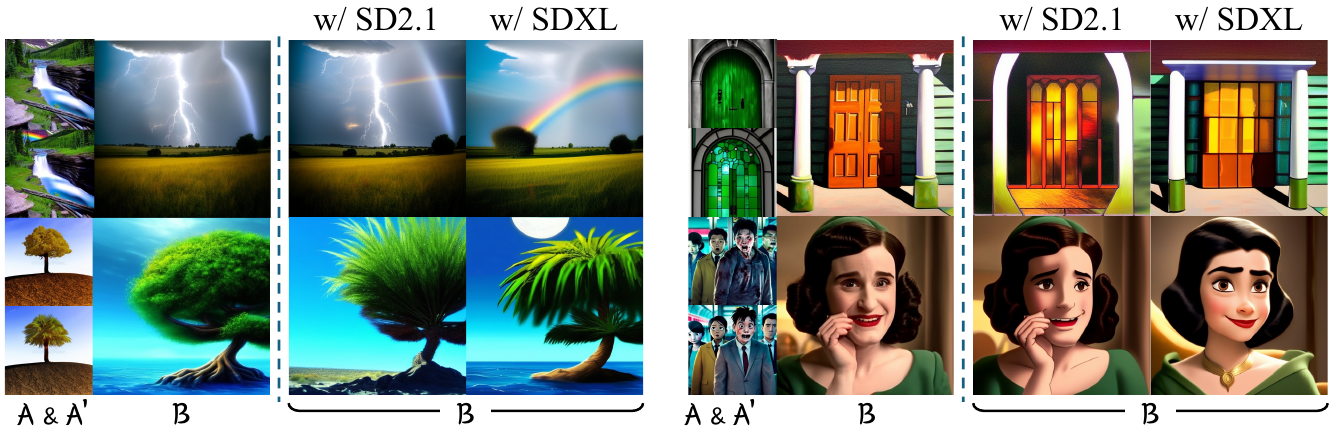}
\caption{\textbf{Transferability to other Stable Diffusion models.} Our Difference Inversion can also be applied to other general Stable Diffusion models, such as SDXL.}
\label{fig:sdxl}
\end{figure}

\paragraph{Transferability.}
Since our Difference Inversion is an inference-based method that does not require access to the pretrained diffusion model weights, it can be applied even if the diffusion model is provided in an API form. Though instruction inversion models such as VISII are also inference-based, they depend solely on specific models (\eg InstructPix2Pix). We test our method on SDXL, which possesses a stronger image prior and is capable of generating more complex and refined images. Fig.~\ref{fig:sdxl} demonstrates the application of Difference Inversion to SDXL.

%% file: sec/6_limitation.tex
\section{Limitations and Discussions}
\label{sec:limitations}

\begin{figure}[t!]
\centering
\includegraphics[width=1\linewidth]{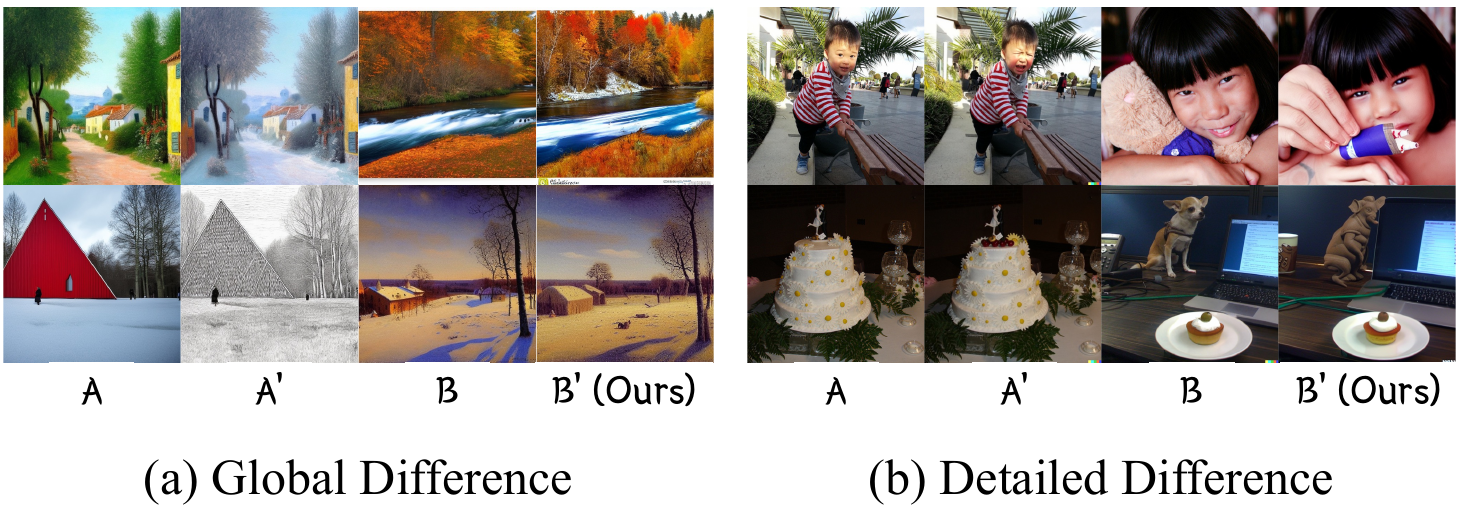}
\caption{\textbf{Failure cases.}
When the difference is complex or difficult to define precisely, Difference Inversion may struggle to capture it effectively.}
\label{failure}
\end{figure}

\paragraph{Failure cases.}
Despite the strong performance of our model across various editing tasks compared to various baselines, there are still some failure cases.
In Fig.~\ref{failure}, it is evident that Difference Inversion struggles to capture global differences (a) or detailed differences (b). For example, global differences (a) such as weather shifts or transformations like image-to-sketch cannot be effectively applied to $B^\prime$ when optimized through text tokens. Similarly, for detailed differences (b), as shown in the top example of changing a boy to a crying boy, the approach may result in changing a girl to another girl or fail to accurately capture the intended crying expression altogether. In cases like adding cherries to a cake (bottom example), such changes may also be poorly reflected in B', highlighting its limitations in capturing intended variations.

\paragraph{Inherent limitations of the CLIP space.}
We hypothesize that the primary reason for the aforementioned failure cases stems from the inherent limitations of the CLIP space. Although we attempt to derive finer Delta using Image and Text Delta within the CLIP space, it does not always align well with human perception, potentially making it unstable for providing precise directional guidance. In the supplementary materials, we illustrate this using a colorization analogy on MNIST data, showing the CLIP similarity for transitions from ground truth $A \rightarrow A^\prime$ and $B \rightarrow B^\prime$. Notably, the CLIP similarity yields only around 0.5, indicating that optimization based on this directional signal is challenging. We anticipate that future approaches may involve using more refined embedding spaces from foundation models such as DINO-v2 or learning additional mapping networks to enhance similarity and stability.

%% file: sec/7_conclusion.tex
\section{Conclusion}
\label{sec:conclusion}
In this paper, we propose Difference Inversion, a model-agnostic inversion method that applies only the differences to $B^\prime$ without introducing unwanted artifacts. Difference Inversion captures precise differences from $A$ and $A^\prime$ and can be applied to any query image $B$. This approach aligns well with the goal of image analogy generation, resulting in the creation of feasible $B^\prime$. Additionally, our method can be applied to various diffusion models, paving the way for new opportunities not only in image editing but also across diverse computer vision tasks using analogy formulations.

%% file: sec/8_ack.tex
\section{Acknowledgements}
\label{sec:ack}
This research was supported by the MSIT(Ministry of Science and ICT), Korea, under the ITRC(Information Technology Research Center) support program(IITP-2024-RS-2023-00258649, 50\%) supervised by the IITP(Institute for Information \& Communications Technology Planning \& Evaluation). This research was also supported by the Institute of Information \&communications Technology Planning \& Evaluation (IITP) grant funded by the Korea government(MSIT) (No. RS-2019-II190079, Artificial Intelligence Graduate School Program(Korea University), 1\%) and the National Research Foundation of Korea(NRF) grant funded by the Korea government(MSIT)(No. RS-2025-00562437 40\%, No. RS-2024-00341514, 9\%)